\documentclass[runningheads]{llncs}
\usepackage{graphicx}
\usepackage{amsmath,amssymb} 
\usepackage{color}
\usepackage{ulem}
\usepackage{subfig}
\usepackage{algorithm}
\usepackage{siunitx}
\usepackage{algpseudocode}
\usepackage[utf8]{inputenc}
\begin{document}
\pagestyle{headings}
\mainmatter

\title{Image Reassembly Combining Deep Learning and Shortest Path Problem} 
\titlerunning{Image Reassembly Combining Deep Learning and Shortest Path Problem}
\authorrunning{M.-M. Paumard \and D. Picard \and H. Tabia}
\author{Marie-Morgane Paumard$^1$ \and David Picard$^{1,2}$ \and Hedi Tabia$^1$\thanks{This work is supported by the Fondation des sciences du patrimoine, LabEx PATRIMA ANR-10-LABX-0094-01}}

\institute{$^1$ETIS, UMR 8051, Université Paris Seine, Université Cergy-Pontoise, ENSEA, CNRS\\
	$^2$Sorbonne Université, CNRS, Laboratoire d'Informatique de Paris 6, F-75005 Paris\\
	\email{ \{marie-morgane.paumard, picard, hedi.tabia\}@ensea.fr}
}

\maketitle

\begin{abstract}
This paper addresses the problem of reassembling images from disjointed fragments.
More specifically, given an unordered set of fragments, we aim at reassembling one or several possibly incomplete images.
The main contributions of this work are: 1) several deep neural architectures to predict the relative position of image fragments that outperform the previous state of the art; 2) casting the reassembly problem into the shortest path in a graph problem for which we provide several construction algorithms depending on available information; 3) a new dataset of images taken from the Metropolitan Museum of Art (MET) dedicated to image reassembly for which we provide a clear setup and a strong baseline.
\end{abstract}

\section{Introduction}
The problem of automatic object reconstruction is very important in computer vision, as it has many potential applications in, e.g.  cultural heritage and archaeology. For instance, given numerous fragments of an art masterpiece, archaeologists may spend a long time searching their correct configuration. 
In recent years, vision-related tasks such as classification~\cite{alexnet}, captioning~\cite{karpathy} or image retrieval~\cite{gordo2016deep} have been tremendously improved thanks to deep neural network architectures, and the automatic reassembly of fragments can also be cast as a vision task and improved using the same deep learning methods.

\begin{figure}
  \centering
  \subfloat[Fragments]{\label{fig:task-a}\includegraphics[scale=0.4]{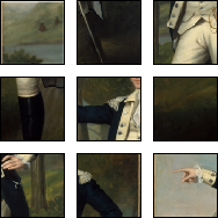}}
  \hspace{30pt}
  \subfloat[Reassembly]{\label{fig:task-b}\includegraphics[scale=0.4]{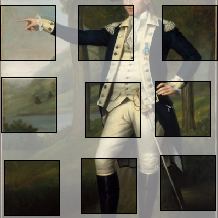}}
  \caption{Example of the reassembly task on the MET dataset}
  \label{fig:task}
\end{figure}

In this paper, we focus on global image reassembly. The fragments are 2D-tiles and the problem consists in finding their approximated position, as shown in Figure \ref{fig:task}.
To solve the problem, we build on the method proposed by Doersch et~al.~\cite{doersch} that proposes to train a classifier able to predict the relative position of a fragment with respect to another one.
We show that solving the reassembly problem from an unordered list of fragments can be expressed as a shortest path problem in a carefully designed graph. The structure of the graph heavily depends on the properties of the puzzle such as its geometry (number of positions and their layout), its completeness (a fragment for each available positions) and its homogeneity (all fragments have a correct position in the puzzle).

Our contributions are the following. First, we propose several deep convolutional neural network architectures for predicting the relative position of a square-cropped fragment with respect to another. The crop allows us to ignore the borders of each piece and to focus on the content in order to achieve a global positioning. Second, we propose several graph construction algorithms that implement the reassembly problem corresponding to the different cases of puzzles depending on the aforementioned properties. Third, we perform extensive experiments of the different neural network and shortest path graph problem combinations on ImageNet~\cite{ILSVRC15} and on a new dataset composed of 14,000 images from the Metropolitan Museum of Art (MET). For this new dataset, we provide a clear setup and evaluation procedure that allows future works on the reassembly problem to be compared.

This paper is organized as follows: in section~\ref{sec:rw}, we present related work on puzzle solving and fragment reassembly as well as relevant literature on feature combination as it is an essential step of the relative position prediction. Next, we detail our propositions for the deep neural network building block and the graph construction algorithms that correspond to the different image reassembly problems. In section~\ref{sec:exp}, we present our experimental setups and analyze the results obtained for different combinations of deep neural networks and graphs.

\section{Related work}
\label{sec:rw}
In this section, we first present the related work on puzzle solving. Then we detail the relevant literature on feature combination.

\subsection{Puzzle solving}
The reconstruction of archaeological pieces of art leads to better understanding of our history and thus attracts numerous researchers, as Rasheed and Nordin described in their surveys \cite{rasheed,survey}. Most publications of this field rely on the border irregularities and aim for precise alignment. They focus on automated reconstruction, such as \cite{mcbride,jampy,zhu} and consider jigsaw puzzle solving with missing fragments or with differently sized tiles \cite{jigsaw1,jigsaw2,jigsaw3}. These methods perform well on a small dataset with only one source of fragments. On the downside, they stall when fragments come from various sources and they require costly human made annotations. Moreover,they are fragile towards erosion and fragment loss.

Without being interested in jigsaw puzzle solving, Doersch et~al. proposed a deep neural network to predict the relative position of two adjacent fragments in \cite{doersch}. The end goal of the authors is to use this task as a pretraining step of deep convolutional neural network (CNN), harnessing the vast amounts of unlabeled images since the ground truth for such task can be automatically generated. The intuitions for training features able to predict their context are the same as what is found in the text literature with word2vec~\cite{mikolov2013distributed} or skip-thought~\cite{kiros2015skip}. In \cite{doersch}, the authors show their proposed task outperforms all other unsupervised pretraining methods. Based on \cite{doersch}, Noroozi and Favaro \cite{noroozi} introduce a network that compares all the nine tiles at the same time. They claim that the complete representation obtained allows discarding the ambiguities that may have been learned with the algorithm proposed by Doersch et~al. Gur et~al. \cite{gur} consider missing fragments, but heavily rely on border to solve the puzzle.

In this paper, we focus on solving the jigsaw puzzle and not on the building of generic images features. In cultural heritage, we have missing pieces, as well as pieces from various images. Therefore, the setup of \cite{noroozi} is impractical, as it requires exactly the nine correct fragments to make a prediction. For this reason, we base our work on the method proposed in~\cite{doersch}, but we do not share the same objective and we bring two significant innovations. First, we consider the correlations between localized parts of the fragments when merging the features, something that is difficult to achieve in \cite{doersch}. We believe these correlations are important, since, e.g., we expect the right part of the baseline fragment to be correlated with the left part of the right fragment. Second, we look for a complete fragment reassembly, which we perform by using the deep neural network predictions to build a shortest path graph problem.

\subsection{Feature combination}

Doersch et~al.~\cite{doersch} separately processed fragments using a deep CNN with shared weights which output comparable features. These features are then serially concatenated and fed to a multi-layer perceptron (MLP) that performs the classification. The full network has been trained in an end-to-end fashion with standard back-propagation using stochastic gradient descent.

In Doersch et~al.~\cite{doersch} formulation, the cross-covariance between the features of both fragments is neglected. Indeed, the output of the CNN can be viewed as localized pattern activations. The prediction of the relative position depends on the conjunction of specific patterns occurring at specific positions in the first fragment and specific patterns occurring at specific positions in the second fragment. It can be argued that a sufficiently deep MLP can model these cross-covariances, but it also seems easier to model them directly.

In \cite{lin2015bilinear}, the authors suggest modeling these co-occurrences of patterns using a bilinear model which can be computed using the Kronecker product of the feature vectors.
They report improved accuracy on fine-grained classification.
However, using the Kronecker product leads to high dimensional features that are intractable in practice.
To overcome this burden, the authors of \cite{gao2016compact} propose to use random projections combined with the Hadamard (element-wise) product to approximate the bilinear model. This strategy is further extended in \cite{hadlr} where the projections are trained in true deep learning fashion. Another factorization based on the Tucker decomposition is also proposed in \cite{mutan} which allows controlling the rank of the considered co-occurrences.

\section{Method}

In this section, we detail our proposed method. We start by presenting the deep CNN model upon which we build to solve the image reassembly problem.

\subsection{Relative position prediction}
To solve a puzzle, we need to pick the fragments to use. We compare each selected fragment with the central fragment and compute their relative position. We examine several ways to articulate this problem.

\subsubsection{Problem formulation}
\begin{figure}[ht]
\centering
\includegraphics[width=1\columnwidth]{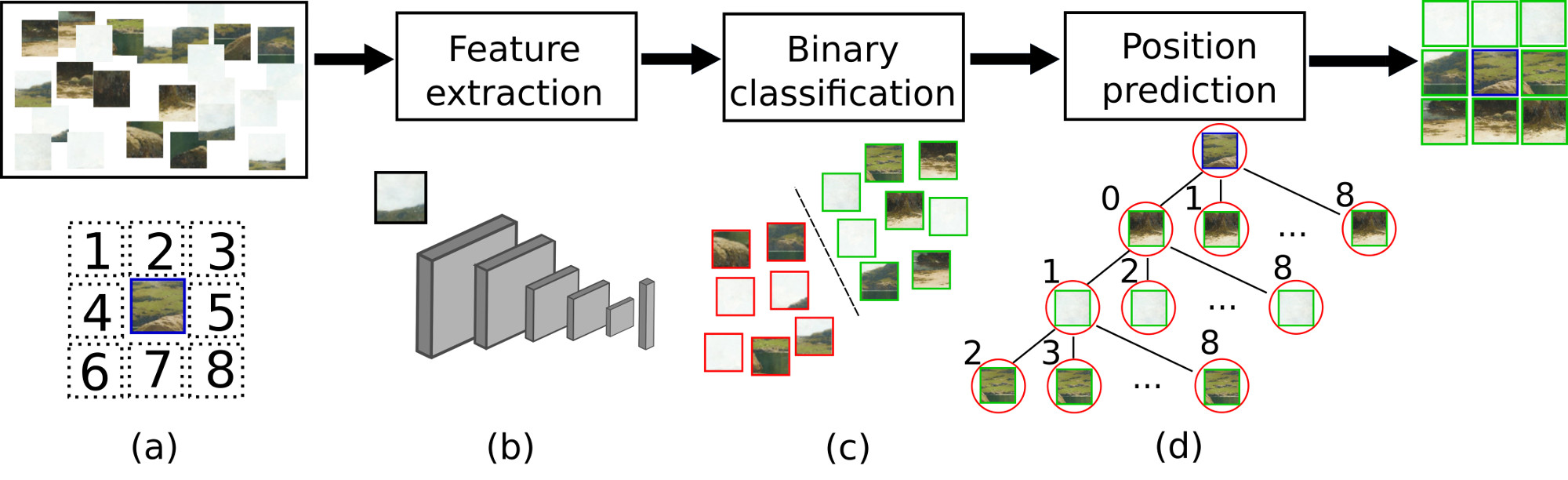}
\caption{\label{fig:overview}Overview of our method. Knowing a central fragment, we are looking for the correct arrangement to reassemble the image (a). We extract the feature of all the fragments (b) and we compare them to the features of the central fragment. We predict which fragments are part of the image (c). We retrieve the top eight fragments and we predict their relative position with respect to the central one. We turn the prediction into a graph (d). We then run a shortest path algorithm to reconstruct the image}
\end{figure}

The first step towards reassembly consists in discriminating between the fragments that may be of use and the others. On our puzzle, it means that we predict which fragments are allegedly extracted from the same image as a given central fragment, which is a binary classification problem. Once only relevant fragments are selected, we model the position prediction as an 8-classes classification problem, as shown in Figure~\ref{fig:overview}. Both these classification tasks are performed by a deep CNN described later.

We also propose an alternative model by merging these two networks into a single network. This single network predicts the relative position of the second fragment among the 8 possible positions and a 9th class, activated if the fragment is not part of the same image.

\subsubsection{Network architecture}

\begin{figure}[ht]
\centering
\includegraphics[width=0.8\columnwidth]{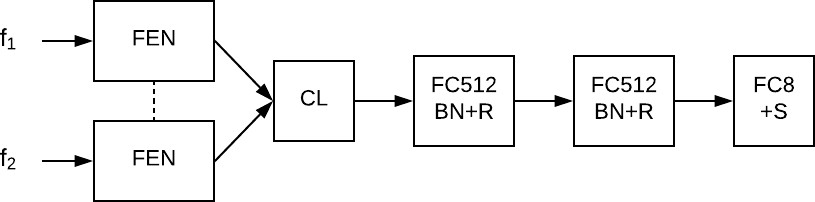}
\caption{\label{fig:net}General network architecture block diagram}
\end{figure}

The global network architecture is described in Figure \ref{fig:net}.
Given two input fragments, we first extract fragment representations using a shared feature extraction network (FEN). We tested the most common architectures and empirically found out that a VGG-like ~\cite{vgg} network works the best. Therefore, the FEN architecture is inspired by a simplified version of VGG~\cite{vgg} and is shown on Table~\ref{tab:archi}. The network is composed of sequences of a $3\times 3$ convolution followed by batch-normalization~\cite{bn}, ReLU activation~\cite{relu} and max-pooling. 
We also tried other models based on more recent architectures such as Resnet~\cite{resnet}, but we empirically found that they were underperforming compared to the simpler architecture. This can be explained by the fact that contrarily to full images, fragments do not contain as much semantic information and thus require less involved features.
Remark also that there is no global pooling \cite{lin2013network} in the FEN and thus spatial information is preserved, which we believe is important for the relative position prediction.

The features of each fragment are then combined in a combination layer (CL). Contrarily to the concatenation that is proposed at this stage in \cite{doersch}, we explore variations on the bilinear product in order to model cross-covariances among the features.
With $\phi_{\text{FEN}}(f)$ the output of the FEN for fragment $f$, the full bilinear product is obtained by using the Kronecker product of the features~\cite{lin2015bilinear}:
\begin{align}
y_{kron} = \phi_{\text{FEN}}(f_1) \otimes \phi_{\text{FEN}}(f_2).
\end{align}

However, this leads to very high dimensional vectors.
Similarly to \cite{hadlr}, we explore a compressed version using the entry-wise product:
\begin{align}
y_{had} = (W^\top\phi_{\text{FEN}}(f_1))\circ (W^\top\phi_{\text{FEN}}(f_2)),
\end{align}
where $\circ$ denotes the Hadamard product.
This compressed version can be efficiently implemented by changing the output size of the last layer in the FEN.

Finally, the classification stage consists of two sequences of a fully connected layer followed by a batch-normalization and a ReLU activation, and a final prediction layer with softmax activation.

\begin{table}[htb]
\begin{center}
  \caption{Architecture of the Feature Extraction Network. Conv: convolution, BN: Batch-Normalization, ReLU: ReLU activation. $OUT$ is chosen among 512, 1024, 2048 and 4096, depending on what merging function we use}
  \label{tab:archi}
  \begin{tabular}{|l|c|c|c|}
    \hline
    \textbf{Layer} & \textbf{Output shape} & \textbf{Parameters shape} & \textbf{Parameters count} \\
    \hline
    Input & $96 \times 96 \times 3$ && 0 \\
    Conv+BN+ReLU & $96 \times 96 \times 32$ & $3 \times 3 \times 32$ & 1k \\
    Maxpooling & $48 \times 48 \times 32$ && - \\
    Conv+BN+ReLU & $48 \times 48 \times 64$ &$3 \times 3 \times 32$& 19k \\
    Maxpooling & $24 \times 24 \times 64$ && -\\
    Conv+BN+ReLU & $24 \times 24 \times 128$ &$3 \times 3 \times 32$& 74k \\
    Maxpooling & $12 \times 12 \times 128$ && - \\
    Conv+BN+ReLU & $12 \times 12 \times 256$ &$3 \times 3 \times 32$& 296k \\
    Maxpooling & $6 \times 6 \times 256$ && -\\
    Conv+BN+ReLU & $6 \times 6 \times 512$ &$3 \times 3 \times 32$& 1.2M \\
    Maxpooling & $3 \times 3 \times 512$ && -\\
    Fully Connected+BN & $OUT$ &&  $OUT_{nb\ param}$ \\
    \hline
  \end{tabular}
\end{center}
\end{table}

\subsection{Puzzle resolution}

Once the position is predicted by the neural network for each fragment, we can solve the puzzle, which consists in assigning fragments to a position in the image. We consider several cases depending on whether we already have a well-positioned fragment, and whether we have supernumerary fragments.

\subsubsection{Problem formulation}

We first consider the case where we are given the central fragment as well as an unordered list of 8 fragments corresponding to the possible neighbors of the central fragment. Solving the puzzle then consists in solving the assignment problem where each fragment $i$ has to be associated with a position $j$. Given the relevance $p_{i,j}$ of fragment $i$ at position $j$, and the assignment variable $x_{i,j} =1$ if fragment $i$ is at position $j$, we want to maximize:

\begin{align}
    \max_{x_{i,j}} \sum_{i,j} p_{i,j} \cdot x_{i,j} \label{eq:one}
\end{align}
under the constraints:

\begin{align}
    \forall j, \sum_{i=0}^8 x_{i,j} = 1 \; \label{eq:two}, \\
    \forall i, \sum_{j=0}^8 x_{i,j} = 1 \; \label{eq:three},\\
    \forall i, j, x_{i,j} \in \{0,1\} \label{eq:four}.
\end{align}
Remark that only one fragment can occupy a position (Equation \ref{eq:two}) and a fragment can be placed only once (Equation \ref{eq:three}).

Then, if we allow the puzzle to be uncompleted (i.e. some positions are not used), we replace the constraint \ref{eq:two} with:
\begin{align}
    \forall j, \sum_{i=0}^8 x_{i,j} \leq 1 \label{eq:five}.
\end{align}

Similarly, if we have supernumerary fragments (i.e. some fragments are not used), we replace the constraint \ref{eq:three} with:
\begin{align}
    \forall i, \sum_{j=0}^N x_{i,j} \leq 1 \label{eq:six}.
\end{align}

Finally, if we do not know which fragment is the central fragment, we have to solve the extended assignment problem where one fragment has to be assigned to the central position and the remaining fragment are assigned to the relative positions. This leads to the following problem:
\begin{align}
    \max_{c,x_{i,j}} \sum_{i,j} p_{i,j,c} \cdot x_{i,j,c}\label{eq:seven}
\end{align}
under the following constraints:
  
\begin{align}
    \nonumber \forall c, j, \sum_{i=0}^N x_{i,j,c} \leq 1;
    \nonumber \forall c, i \neq c, \sum_{j=0}^8 x_{i,j,c} \leq 1;
    \nonumber \forall c, j, \forall i \neq c,  x_{i,j,c} \in \{0,1\};\\
    \nonumber \forall c, j, \forall i=c, x_{i,j,c}=0.  \label{eq:ten}
\end{align}

\subsection{Graph formulation}

Solving the mentioned problem can be done by finding the shortest path in a corresponding directed graph, which can be done using Dijkstra's algorithm or any of its variants. In this section, we show how to construct such graphs.

Each graph starts with a source $S$ and ends with a sink $T$. Each subsequent depth level from $S$ corresponds to a fragment. All nodes at a given depth $i$ from $S$ correspond to the position that could be assigned to a fragment $i$ given all previous assignments. Each edge receives the corresponding classification score as the weight.

\begin{figure}
\centering
\subfloat[]{
\includegraphics[width=0.2\columnwidth]{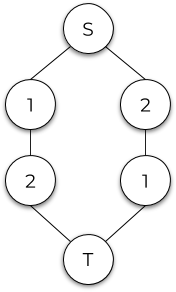}
\label{fig:graph-8}
}
\subfloat[]{
\includegraphics[width=0.46\columnwidth]{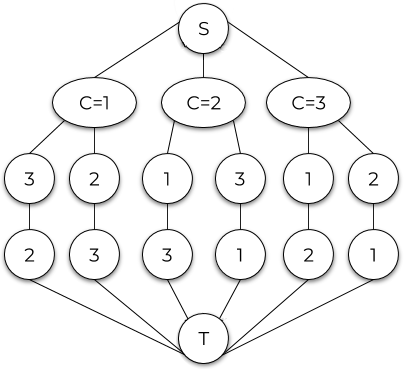}
\label{fig:graph-9}
}
\caption{Examples of graphs for a complete problem with known and unknown central fragment, for empty 2 positions}
\label{fig:graphs}
\end{figure}

When the central fragment is known and we have the exact number of missing fragments, the construction procedure is given in Algorithm \ref{algo:graph1}. We also give a very simple example with only two relative positions in Figure \ref{fig:graph-8}.

\begin{algorithm}
\caption{Graph building from central fragment}
\label{algo:graph1}

\begin{algorithmic}[1]
\Procedure{Construct\_edges}{$Y$}\Comment{Y is the predicted values matrix for $i,j$}
\State $empty\_pos \gets [1..9]$
\State $used\_pos \gets [S]$
\State $next\_frag \gets 1$
\State $tree \gets$ \textsc{Add children}($Y, empty\_pos, used\_pos, next\_frag$)
\State \textbf{return} $tree$\Comment{The list of the edges: related fragment, position of the previous node, position of the current node, cost of the edge.}
\EndProcedure
\end{algorithmic}

\begin{algorithmic}[1]
\Procedure{Add\_children}{$Y, empty\_pos, used\_pos, next\_frag$}
\State $edges \gets [~]$

\If{$empty\_pos$ is empty} 
\State $edges \gets [(None, last(used\_pos), T, 0)]$ \Comment{Append the $j\rightarrow T$ edge}
\State \textbf{return} $edges$
\EndIf

\For{$pos$ in $empty\_pos$}

\State $edges \gets edges \cup [(next\_frag, last(used\_pos), pos, Y[next\_fragment, pos])]$
\State $empty\_pos \gets empty\_pos \setminus pos$
\State $used\_pos \gets used\_pos \cup pos$
\State $edges \gets edges \cup (\textsc{Add children}(Y, empty\_pos, used\_pos, next\_frag+1$))
\EndFor

\State \textbf{return} $edges$
\EndProcedure
\end{algorithmic}
\end{algorithm}

In the case where the central fragment is known, the size of the resulting graph is $|E| = \frac{n!}{(n-p)!}+\sum_{i=n-p}^{n-1}\frac{n!}{i!}$ for the number of edges and $|N| = 2+\sum_{i=n-p}^{n-1}\frac{n!}{i!}$ for the number of vertices, with $n$ the number of fragments and $p$ the number of positions. With 8 fragments and positions, this corresponds to $|E| = 150\si{k}$ and $|N| = 100\si{k}$.

In the case where we do not know the central fragment, we simply perform the central fragment selection as a first step. The first expansion from S consists in all the possible cases where each fragment is used as the central fragment. The corresponding subgraphs are the built using Algorithm \ref{algo:graph1}. The size of the resulting graph is unchanged, except we have $n+1$ fragments, with $n$ the number of the fragment to be assigned to a relative position. With $n=8$, we obtain $|N| = 1\si{M}$ and $|E| = 1.3\si{M}$. We show in Figure \ref{fig:graph-9} a simplified example with 3 fragments and 2 relative positions.

\begin{figure}
\centering
\includegraphics[width=0.45\columnwidth]{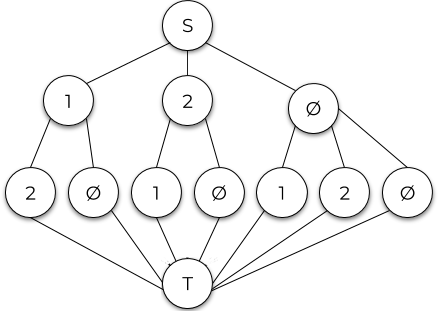}
\caption{Example of a graph allowing empty positions}
\label{fig:graph-incomplete}
\end{figure}

Finally, we now consider the case where the puzzle may not be solved with all the fragments we have. This means that we can have more than 8 fragments, coming from various sources. We also may have missing fragments, and consequently, we prefer an algorithm that proposes an incomplete solution than a wrong reassembly. We construct a graph allowing such configurations by enabling the algorithm to pick no fragment. A simplified example of the graph shown in Figure \ref{fig:graph-incomplete}. 

The graph building algorithm is similar to the Algorithm \ref{algo:graph1}; if we add a position \O ~at the $antecedents$ list, we do not exclude it from the further available choices, as detailed in Algorithm \ref{algo:graph2}.
This graph has:

\begin{align}
|N| = 2+\sum_{l=0}^n\sum_{k=p-l}^p\dbinom{l}{p-k}\frac{(k+1)p!}{k!}
\end{align}
vertices and
\begin{align}
    |E| = \sum_{k=p-n}^p\dbinom{l}{p-k}\frac{(k+1)p!}{k!}+\sum_{l=0}^n\sum_{k=p-l}^p\dbinom{l}{p-k}\frac{(k+1)p!}{k!}
\end{align}
edges, with $n$ fragments and $p$ positions. If the breadth of the graph is limited by the number of position, the depth depends on the number of fragments. In the case of 10 fragments and 8 relative positions, the size of the graph is $|E|=5\cdot10^9$ and $|N|=4\cdot10^8$.

Once the graph has been set up, the shortest path from $S$ to $T$ can be found with Dijkstra's algorithm \cite{dijkstra} for which the complexity is $\mathcal{O}(|E|+|N|\times \log(N))$.

\begin{algorithm}
\caption{Graph building with empty positions}
\label{algo:graph2}

\begin{algorithmic}[1]
\Procedure{Add\_children}{$Y, empty\_pos, used\_pos, next\_frag$}
\State $edges \gets [~]$

\If{$empty\_pos$ is empty or $next\_frag>n$} 
\State $edges \gets [(None, last(used\_pos), T, 0)]$ \Comment{Append the $j\rightarrow T$ edge}
\State \textbf{return} $edges$
\EndIf

\For{$pos$ in $empty\_pos \cup $\O}

\State $edges \gets edges \cup [(next\_frag, last(used\_pos), pos, Y[next\_fragment, pos])]$
\If{$pos$ in $empty\_pos$}
\State $empty\_pos \gets empty\_pos \setminus pos$
\EndIf
\State $used\_pos \gets used\_pos \cup pos$
\State $edges \gets edges \cup (\textsc{Add children}(Y, empty\_pos, used\_pos, next\_frag+1$))
\EndFor

\State \textbf{return} $edges$
\EndProcedure
\end{algorithmic}
\end{algorithm}

\subsubsection{Greedy method}
We implement a greedy method to enable us to benchmark the Dijkstra algorithm. We solve iteratively the puzzle, picking at each step the top value from the neural network predictions. We expect this method will make worst choices than Dijkstra'sn considering the dependencies between the steps.

\section{Experiments}
\label{sec:exp}

In this section, we first describe our experimental setup as well as our new dataset related to cultural heritage. Then, we give experimental results on the classification task and on full image reassembly.

\subsection{Experimental setup}

The neural networks are trained using fragments from 1.2M images of ImageNet. We use 50k images to evaluate the classification accuracy. Each image is resized and square-cropped to $398 \times 398$ pixels, and divided into 9 parts separated by a 48-pixel margin, corresponding to the erosion of the fragments. Each fragment has a size of $96 \times 96$ pixels and has to be contained in one of the 9 parts, which means that it can be chosen within a $\pm 7$-pixels range in each direction.

For the reassembly, the neural networks are then fine-tuned on a cultural heritage dataset, consisting of 14,000 open-source images from the Metropolitan Museum of Art. Such dataset is close to our aimed application, puzzle solving for cultural heritage

\subsection{Classification}

To evaluate our proposed architectures for classification, we reproduce the architecture Doersch et~al. detailed in \cite{doersch}. The authors reported a 40\% accuracy on ImageNet for the 8-classes classification task. Replicating the architecture of their neural network, we obtain an accuracy of 57\%. This may be explained by the tuning of the hyperparameters. 

In Table \ref{tab:merge}, we report the accuracy for the different combination layers on the 8-classes problem on ImageNet validation images.
As we can see, the Kronecker product obtain slightly better results than the concatenation. However, using the low-rank approximation of \cite{hadlr} yields lower results which means that the full covariances are needed to obtain the best performances. Remark that all of our architecture outperforms the architecture proposed in \cite{doersch}.

\begin{table}
\begin{center}
\caption{Accuracy for different fusion strategies, for the 8-classes classification problem on ImageNet validation. $^\star$ denotes our implementation}
\label{tab:merge}
\begin{tabular}{l|r}
\textbf{Fusion} & \textbf{Accuracy} \\
\hline
Doersch et~al. \cite{doersch}$^\star$ & 57.0\% \\
Concatenation & 64.6\% \\
Kronecker product & 66.4\% \\
Hadamard product & 59.2\% \\
\end{tabular}
\end{center}
\end{table}

We show the results of the sequential classification approach (2 classes, then 8 classes) and the joint classification approach (9 classes)  on Table \ref{tab:classes}.
For the binary classification problem, we set the proportion of fragments belonging to the same image to 50\% and we obtain 92.5\% accuracy. which means that deciding whether two fragments belong to the same image seems to be an easy problem.
For the 8 classes problem, we obtain 66.4\% accuracy. It is not surprising to reach around 33\% error since many fragments are ambiguous with respect to the precise location among three positions. For example, sky fragments are easy to classify on top with respect to the central fragment, but which of the three top positions is often difficult to guess.
Finally, the joint classification problem achieves 64.2\% (the proportion of fragment belonging to the same image was set to 70\%), which indicates that solving the joint problem is not harder than solving the sequence of simpler problems.

\begin{table}
\begin{center}
\caption{Classification accuracy for the 2-classes, 8-classes and 9-classes problems on Imagenet, using the Kronecker combination layer}
\label{tab:classes}
\begin{tabular}{l|r}
\textbf{Problem} & \textbf{Accuracy} \\
\hline
2-classes neighborhood classifier & 92.5\% \\
8-classes position classifier & 66.4\% \\
9-classes classifier & 64.2\%
\end{tabular}
\end{center}
\end{table}

\subsection{Reassembly}
In Table \ref{tab:dij-8}, we compare various cases of reassembly tasks, using two different accuracy measures. The reconstruction accuracy describe if the puzzle is perfectly solved. The position accuracy counts how many fragments are well placed.

\begin{table}
\begin{center}
\caption{Reconstruction accuracies and position accuracies for different reassembly problems}
\label{tab:dij-8}
\begin{tabular}{l|c|c|c|c}
& \multicolumn{2}{c|}{Reconstruction accuracy} & \multicolumn{2}{c}{Position accuracy} \\
Problem & Greedy & Dijkstra & Greedy &  Dijkstra\\
\hline
Central known, complete puzzle & 41.0 & 44.4 & 87.7 & 89.9 \\
Central unknown, complete puzzle & 36.2 & 39.2 & 69.5 & 71.1 \\
Central known, incomplete puzzle & 26.5 & 29.5 & 80.5 & 82.4
\end{tabular}
\end{center}
\end{table}

As we can see, in the case of the complete puzzle where the central fragment is known, we are able to perfectly reassemble the image in 44.4\% of the cases using Dijkstra's algorithm, which represent a 3\% improvement over the greedy algorithm, which is closer to the optimal solution than one might think. Remark that the position accuracy is around 90\%, which is much better than the 66.4\% accuracy of the neural network used to solve the task. This shows that solving the reassembly problem can remove some uncertainty the classifier has.

When the central fragment is not know, the reassembly accuracy drops only to 39.2\% and the position accuracy drops to 71.1\%. This means that reassembling the image without knowing the central fragment is not much more complicated than with the central fragment known, however, if that first step is missed, then all subsequent assignments are likely to be wrong.

We consider adding outsider fragments to the puzzle (Table \ref{tab:additional}), making the accuracy drop. The increase of computation time triggered by the addition is reasonable as long as the puzzle still contains 9 pieces. Any increment of the number of pieces leads to an factorial increase of the number of solution, making the problem quickly intractable. Nonetheless, any puzzle can be divided into $3\times3$ puzzles, that would be solved individually and fused.

\begin{table}
\begin{center}
\caption{Position and reconstruction accuracies with additional fragments}
\label{tab:additional}
\begin{tabular}{c|c|c|c}
Number of additional fragments & 0 & 1 & 2\\
\hline
Reassembly accuracy (Dijkstra) & $44.4\%$ & $26.3\%$ & $14.3\%$\\
Position accuracy (Dijkstra) & $89.9\%$ & $75.3\%$ & $64.8\%$
\end{tabular}
\end{center}
\end{table}

\begin{figure}
\centering
\subfloat{\includegraphics[width=0.25\columnwidth]{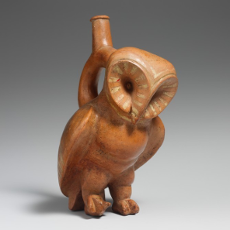}}
\subfloat{\includegraphics[width=0.25\columnwidth]{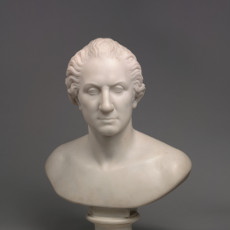}}
\subfloat{\includegraphics[width=0.25\columnwidth]{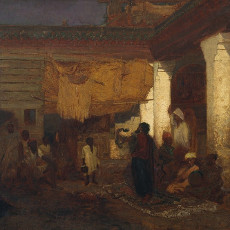}}

\subfloat{\includegraphics[width=0.25\columnwidth]{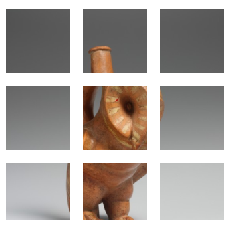}}
\subfloat{\includegraphics[width=0.25\columnwidth]{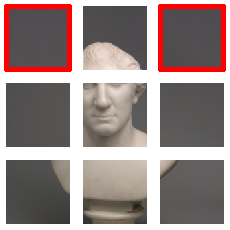}}
\subfloat{\includegraphics[width=0.25\columnwidth]{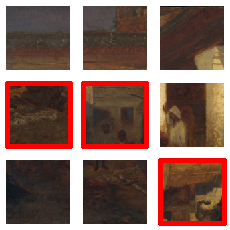}}

\caption{Examples of reconstructions with unknown central fragment. The red outlined fragments are misplaced}
\label{fig:example1}
\end{figure}

In Figure \ref{fig:example1}, we selected few reconstructions with unknown central fragment. The two first images illustrate a significant part of our dataset in which it is easy to misplace background fragments. Most of our reconstruction errors are due to similar reversals. The type of error illustrated by the right image is rare; but, when the central fragment is misplaced, all the other fragments are shifted.

\begin{figure}
\centering
\subfloat{\includegraphics[width=0.25\columnwidth]{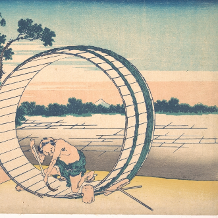}}
\subfloat{\includegraphics[width=0.25\columnwidth]{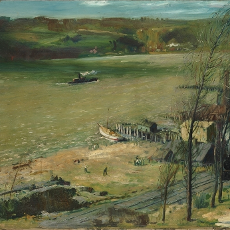}}
\subfloat{\includegraphics[width=0.25\columnwidth]{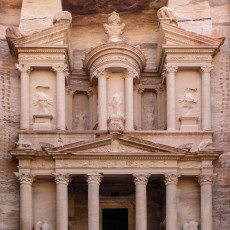}}

\subfloat{\includegraphics[width=0.25\columnwidth]{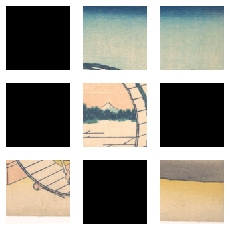}}
\subfloat{\includegraphics[width=0.25\columnwidth]{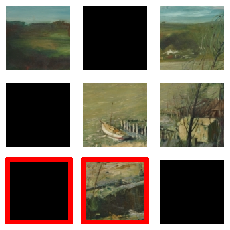}}
\subfloat{\includegraphics[width=0.25\columnwidth]{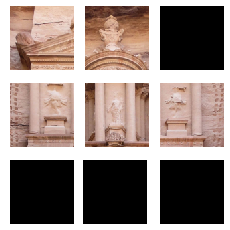}}

\caption{Examples of reconstructions with 4 missing fragments. The red outlined fragments are misplaced}
\label{fig:example2}
\end{figure}

Finally, we study the case where we have missing fragments (Table \ref{tab:dij-8}, last row). In that scenario, only 4 fragment are taken from the image while 8 positions are available. We are still able to predict the position with high accuracy (surprisingly better than in the case where the central fragment is unknown), but perfectly reassembling the image is very difficult. This means that the algorithm tends to drop fragments instead of assigning them to an uncertain location. Figure \ref{fig:example2} shows examples of reconstructions in the case of missing fragments.

\section{Conclusion}
\label{sec:conc}

In this paper, we tackled the image reassembly problem where given a unordered list of image fragments, we want to recover the original image.
To that end, we proposed a deep neural network architecture that predicts the relative position of a given pair of fragments.
Then, we cast the reassembly problem into a shortest path in a graph algorithm for which we propose several construction algorithms depending on whether the puzzle is complete or if there are missing pieces.
We propose a new dataset containing 14,000 images to test several reassembly tasks and we show that we are able to perfectly reassemble the image 44.4\% of the time in the simpler case and 29.5\% of the time if there are missing pieces.

\bibliographystyle{splncs}
\bibliography{egbib}
\end{document}